\documentclass[sigconf]{acmart}
\usepackage{adjustbox}
\usepackage{booktabs}
\usepackage{graphicx}
\usepackage{subcaption}
\usepackage{multirow}
\usepackage{array}
\usepackage{amsmath}
\usepackage{enumitem}
\usepackage{algorithm}
\usepackage{algpseudocode}

\usepackage{tabularx, booktabs}
\usepackage{pgfplots}
\usepackage{tikz}
\usepackage{xcolor}
\usepackage{graphicx}
\usepackage{multirow}
\usepackage{pifont}
\usepackage{algpseudocode}
\usepackage{verbatim}

\pgfplotsset{compat=1.11,
    /pgfplots/ybar legend/.style={
    /pgfplots/legend image code/.code={%
       \draw[##1,/tikz/.cd,yshift=-0.25em]
        (0cm,0cm) rectangle (3pt,0.8em);},
   },
}
\definecolor{applegreen}{rgb}{0.55,0.71,0.0}
\AtBeginDocument{%
  }

\copyrightyear{2025}
\acmYear{2025}
\setcopyright{acmlicensed}
\acmConference[WSDM '25] {Proceedings of the Eighteenth ACM International Conference on Web Search and Data Mining}{March 10--14, 2025}{Hannover, Germany.}
\acmBooktitle{Proceedings of the Eighteenth ACM International Conference on Web Search and Data Mining (WSDM '25), March 10--14, 2025, Hannover, Germany}
\acmISBN{979-8-4007-1329-3/25/03}
\acmDOI{10.1145/3701551.3703586}

\begin{document}

\title{UniGLM: Training One Unified Language Model for \\ Text-Attributed Graph Embedding}

\author{Yi Fang}
\orcid{0009-0002-9048-0281}
\affiliation{%
\department{SFSC of AI and DL}
\institution{New York University Shanghai}
\country{}
}
\email{yf2722@nyu.edu}

\author{Dongzhe Fan}
\orcid{0009-0002-3180-2156}
\affiliation{
\department{SFSC of AI and DL}
\institution{New York University Shanghai}
\country{}
}
\email{df2362@nyu.edu}

\author{Sirui Ding}
\orcid{0000-0001-5827-8214}
\affiliation{
  \department{Bakar Computational Health Sciences Institute}
  \institution{University of California San Francisco}
  \country{}
}
\email{siruidingdavid@gmail.com}

\author{Ninghao Liu}
\orcid{0000-0002-9170-2424}
\affiliation{
  \department{School of Computing}
  \institution{University of Georgia}
  \country{}
}
\email{ninghao.liu@uga.edu}

\author{Qiaoyu Tan}
\authornote{Corresponding author.}
\orcid{0000-0001-8999-968X}
\affiliation{%
  \department{SFSC of AI and DL}
  \institution{New York University Shanghai}
  \country{}
}
\email{qiaoyu.tan@nyu.edu}

\renewcommand{\shortauthors}{Trovato et al.}

\begin{abstract}
Representation learning on text-attributed graphs (TAGs), where nodes are associated with textual descriptions, is crucial for textual and relational knowledge systems, such as social media and recommendation scenarios. However, state-of-the-art embedding methods for TAGs primarily focus on fine-tuning pre-trained language models (PLMs) using structure-aware training objectives. While effective, these methods are tailored for individual TAG and cannot generalize across various graph scenarios. Given the shared textual space, leveraging multiple TAGs for joint fine-tuning, aligning text and graph structure from different aspects, would be more beneficial. Therefore, we propose the Unified Graph Language Model (\textbf{UniGLM}), a novel foundation model pretrained over multiple TAGs from a variety of domains, which can generalize well to both in-domain and cross-domain graph scenarios. Specifically, UniGLM fine-tunes well-established PLMs (e.g., Sentence-BERT) using a domain-aware contrastive learning objective that unifies structure heterogeneity and node statistics across various domains with an adaptive and learnable positive sample selection scheme. Additionally, a lazy updating module is introduced to speed up training by reducing repetitive encoding of positive samples. Extensive datasets across multiple domains, downstream tasks (node classification and link prediction), and a spectrum of graph backbones (supervised and self-supervised graph models) are conducted to compare UniGLM with state-of-the-art baselines. Our empirical observations suggest that UniGLM can generate informative representations for cross-domain graphs observed in the training. More importantly, UniGLM also exhibits competitive transfer ability in encoding unseen TAGs that are not used for training. This study provides deep insights into how to adapt PLMs to graph data and demonstrates the potential of building foundation model for graph representation learning. 

\end{abstract}


\maketitle

\section{Introduction}
\label{sec:intro}
Text-attributed graphs (TAGs) have been widely adopted to represent complex relationships between textual entities in real-world textual and relational knowledge systems, including social media, recommendation systems, and knowledge base. Unlike standard graphs, nodes in TAGs are represented by text attributes. A typical example is academic citation network, where nodes represent scientific papers and edges indicate citations. To learn from TAGs, graph embedding (GE)~\cite{perozzi2014deepwalk,grover2016node2vec,zhang2018network,wu2020comprehensive}, which maps nodes into embedding vectors that preserve both textual and structure information, has recently garnered significant attention. 

Prior GE studies~\cite{he2024harnessing,chen2024exploring} on TAGs primarily focus on two stages: text transformation and graph structure modeling. In the first stage, text attributes are transformed into numerical feature vectors via shallow embedding models such as Word2vec~\cite{mikolov2013efficient} and Bag-of-Words (BoW)~\cite{harris1954distributional}. Subsequently, the transformed node features, along with graph structure, are often fed into graph neural networks (GNNs)~\cite{kipf2016semi,zhou2020graph} for representation learning, where the objective is to either (1) reconstruct the observed links~\cite{kipf2016variational,li2023s,tan2023s2gae} or transformed features~\cite{hou2022graphmae,shi2023gigamae,hou2023graphmae2}, or (2) learn invariant representations between graph augmentations~\cite{you2020graph,zhang2024graph,zhu2021graph,xu2021infogcl,thakoorlarge}, i.e., graph contrastive learning. Despite their promising results, they may be suboptimal in effectively integrating text semantics and structure knowledge.

In recent years, there has been a notable shift of interest in graph embedding from shallow models to pre-trained language models (PLMs) such as {BERT}~\cite{devlin2019bert} and Sentence-Bert~\cite{reimers2019sentence}. The high-level idea is to jointly learn text knowledge and graph structure within a single encoder, either by developing nested graph-BERT architectures~\cite{yang2021graphformers} or by designing structure-aware training signals~\cite{chien2021node,jin2023patton}. Despite their popularity, these methods face limitations in generalization capability because they fine-tune the PLM model for a single particular TAG, making it ineffective for transferring to other TAGs for graph embedding. Given that text attributes provide a unified semantic space across different TAGs, leveraging multiple TAGs for a joint fine-tuning is a promising yet under-explored research direction, supported by the scaling law~\cite{kaplan2020scaling}.

However, training a unified PLM model for multiple TAGs presents several challenges. \textbf{Firstly}, extracting effective structural information across various graph scenarios while maintaining their unique statistics for LM fine-tuning is difficult. Given the diversity and variability of TAGs, local structures such as node degrees and global structures within the graph vary from nodes to nodes and graphs to graphs.  \textbf{Secondly}, directly combining multiple TAGs for joint PLM training may suffer from memory and training efficiency issues due to the non-i.i.d. nature of graphs. 
Unlike pure text-based LM training, textual nodes in TAGs are strongly correlated with each other. Consequently, anchor nodes and their structurally similar neighbors need to be processed by PLM simultaneously, leading to significant trade-offs in computational and memory consumption.

To address the aforementioned challenges, we propose a novel unified graph language model (UniGLM) framework, a pure PLM-based graph embedding approach tailored for multiple TAGs. The key insight is to enhance PLM's (e.g., Sentence-BERT) graph embedding capability by fine-tuning it using large-scale, diverse, and cross-domain text-to-structure knowledge based on \textit{domain-aware contrastive learning}. Specifically, to tackle the first challenge, we introduce an adaptive and learnable positive sample selection technique that identifies positive samples of an anchor node considering its local, global, and graph-specific contexts. This sampling strategy is personalized and can effectively align textual nodes and their important neighbors well across different TAGs. After that, the PLM model is optimized by maximizing mutual information between the anchor node representation and the most informative positive sample adaptively selected from the candidate pool in a learnable fashion. This learnable positive generation scheme further enhances the learning capacity of UniGLM by focusing on positive samples that provide sufficient permutations to contrast anchor nodes. To address the second challenge, we devise a dynamic memory bank to encode positive samples off-the-fly, thereby accelerating the training speed by avoiding repetitive encoding of positive samples' text attributes via PLM. 
The key contributions of our work are summarized as follows: 

\begin{itemize}[leftmargin=*]
\item We explore the development of a generalist embedding model for TAGs and introduce UniGLM, a novel language model pre-training framework tailored for a set of TAGs. To the best of our knowledge, UniGLM is the first graph embedding foundation model for TAGs. 
\item We propose an adaptive and learnable positive sample selection method for sampling positive samples of each node for domain-aware contrastive learning. Unlike standard sampling strategies, our personalized scheme identifies positive samples based on nodes' local, global, and graph-related contexts, thereby unifying graph structures across various TAGs. 
\item Subsequently, a learnable positive generation module is introduced to adaptively select positive samples from the pool aimed at enhancing the contrastive objective by focusing on informative positives (a.k.a. effective node augmentation). Besides, we devise a simple yet effective dynamic embedding table scheme to encode sampled positive samples off-the-fly, accelerating the training process by using historical knowledge as supervision. 
\item We conducted extensive experiments on 9 benchmark TAGs of varying sizes and domains. Empirical results show that UniGLM not only outperforms state-of-the-art graph embedding models across various downstream tasks (node classification and link prediction) and backbones (GNNs and MLPs), but also can generate informative embeddings for unseen TAGs.
\end{itemize}

\section{Related Work}
\label{related-work}
\subsection{Representation Learning on TAGs}
Text-attributed graphs (TAGs) have become a pivotal area of study due to their ability to encapsulate complex relationships between textual data and graph structures. Initially, TAGs employed shallow embedding techniques that were limited in their capacity to meld textual content robustly with graph topology. With the advent of pre-trained language models (PLMs), a more sophisticated extraction of text features became possible. These features are then integrated into graph neural networks (GNNs). However, challenges persist as the integration often leads to suboptimal performance. To address these issues, recent methodologies have sought to more effectively amalgamate PLM features with GNN architectures, thereby enhancing overall model effectiveness. Notably, LM-GNN\cite{ioannidis2022efficient} has pioneered the joint training of large-scale language models with GNNs, creating a unified learning framework. GIANT\cite{chien2021node} leverages extreme multi-label classification to fine-tune language models using graph-based data. Furthermore, GLEM\cite{zhao2023learning} introduces an innovative iterative update mechanism between LM and GNN components in its Expectation-Maximization framework. Another significant advancement is Patton\cite{jin2023patton}, which introduces a graph-aware pre-training approach that includes network-contextualized masked language modeling and masked node prediction, specifically designed to enhance the relevance of pre-trained embeddings for graph-oriented tasks.
 
\subsection{Graph Embedding Foundation Models}
Graph Foundation Models (GFMs) are designed to operate across a diverse range of graphs and tasks, marking a significant step toward universal graph embedding. OpenGraph\cite{xia2023opengraph}, proposed by Xia et al., excels in zero-shot learning scenarios, illustrating the model's ability to generalize well without tailored training. Self-supervised learning (SSL) techniques play a critical role in the pre-training of GFMs, with Zhao et al.\cite{zhao2023selfsupervised} classifying SSL tasks to better harness graph-embedded knowledge, thus enhancing model adaptability. Liu et al.\cite{liu2023towards} discuss the ongoing challenges and potential future directions for GFMs, emphasizing the need for continued innovation. Meanwhile, Tan et al.\cite{tan2023s2gae} introduce an advanced structure reconstruction approach in their model, aiming to further boost the generalizability of GFMs. For those interested in a deeper dive into Graph Foundation Models, particularly regarding their generalization capabilities, scalability, and efficiency, the comprehensive analysis provided in the paper by Xu et al.\cite{xu2024graphfm} is recommended. This work not only discusses various self-supervised GNN models but also includes experimental code, facilitating hands-on exploration and further research in this rapidly evolving field.

\subsection{Graph-aware Large Language Models}
Beyond the typical encoder-only configurations, decoder-only large language models (LLMs) are now being adapted for graph-related tasks. GraphGPT~\cite{tang2024graphgpt} employs a novel projector mechanism to align graph tokens with a pre-trained vocabulary through instruction tuning, facilitating better model understanding and generation of graph structures. LLaGA~\cite{chen2024llaga} melds the capabilities of large language models (LLMs) with the unique requirements of graph-structured data, preserving the versatility of LLMs while reformulating graph data to be more amenable to standard LLM inputs. Additionally, MolecularGPT\cite{liu2024moleculargpt} showcases the potential of LLMs in acting as efficient few-shot predictors for molecular properties, highlighting the model’s adaptability to specialized domains.

\section{Preliminary}
\begin{figure*}[h]
\centering
\includegraphics[width=15cm]{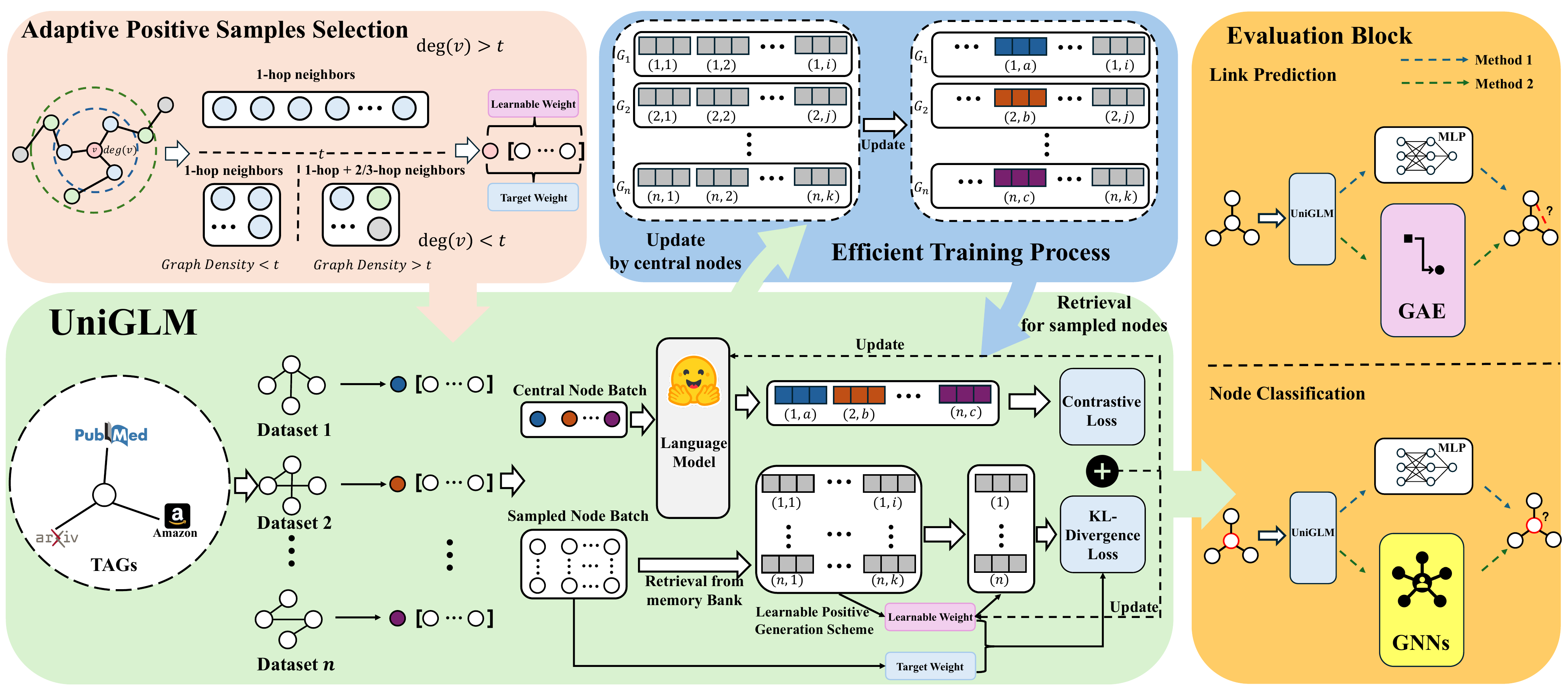}
\caption{The proposed UniGLM framework. The UniGLM framework trains a unified graph encoder across multiple TAGs using domain-aware contrastive learning, instead of learning separate language models for each TAG. To ensure effective and efficient textual-to-structure alignment, we introduce an adaptive and learnable positive sample selection scheme and a lazy updating strategy. UniGLM serves as a foundational embedding model for TAGs, consistently delivering strong performance across various downstream tasks and backbones.
}
\label{fig:uniglm}
\end{figure*}
In this section, we introduce the notation, research problem definition, and the motivation behind learning from multiple TAGs. 

\noindent\textbf{Notations and Problem Definition.} We are given $m$ TAGs, denoted as $\{G_i|i=1,2,...,m\}$, where $G_i=(\mathcal{V}, \mathcal{T}, \mathbf{A})$ represents the $i$-th TAG with $n_i$ nodes. $\mathcal{V}$ is the set of nodes and $\mathbf{A} \in \mathbb{R}^{n_i \times n_i}$ is the adjacency matrix. Each node $v \in \mathcal{V}$ is associated with a textual attribute $\mathcal{T}_v$, and $\mathcal{T} = \{\mathcal{T}_v \mid v \in \mathcal{V}\}$ denotes the attribute set.

\noindent\textbf{TAG Embedding.} 
Given a TAG $G_i=(\mathcal{V},\mathcal{T},\mathbf{A})$, a standard embedding model aims to learn a graph encoder $f_i$ that maps nodes in $G_i$ into embedding vectors, preserving both textual ($\mathcal{T}$) and structure ($\mathbf{A}$) knowledge. Therefore, for $m$ TAGs, traditional methods will learn independent $m$ graph encoders, denoted as $\{f_i\}_{i=1}^m$.

\noindent\textbf{\underline{Motivation}.} Learning one graph encoder for each particular TAG is the de facto standard in state-of-the-art graph embedding literature. However, we argue that this setting is suboptimal for two main reasons. \textbf{\textit{i)} Deployment inefficiency}. As discussed above, this learning procedure requires the development of $m$ separate graph encoders for all TAGs, significantly increasing the deployment and maintenance costs in practice. \textbf{\textit{ii)} Limited performance}. Given the shared textual space across various TAGs, pre-training a language model for a single TAG is inherently less effective because it cannot leverage the text-to-structural knowledge across TAGs. According to the scaling law~\cite{kaplan2020scaling}, incorporating more structure-aware textual knowledge for collaborative language model fine-tuning may be advantageous. Motivated by this, we aim to explore learning from multiple TAGs, as follows. 

\noindent\textbf{Learning on Multiple TAGs.} Given a set of TAGs $\{G_i|i=1,2,...,m\}$, our objective is to develop a single unified graph encoder $f$, such that the textual-to-structure knowledge across $m$ TAGs is collectively preserved within the embedding space.


\section{Methodology}
In this section, we present the details of UniGLM, as depicted in Figure~\ref{fig:uniglm}. First, we introduce the contrastive-based collective language model pre-training pipeline in Section~\ref{sec-training}. Then, in Section~\ref{sec-samples},  we elaborate on the methodology to adaptively select positive samples for collaborative training. After that, we introduce how to aggregate positive samples in Section~\ref{sec-aggregation}. Finally, in Section~\ref{sec-memory-efficiency}, we introduce a simple but effective optimization strategy (Embedding Table) to accelerate our learning process. 

\subsection{The Overall Pipeline of Collaborative Language Model Pre-training}
\label{sec-training}
Learning from multiple TAGs is challenging due to the heterogeneity of textual attributes and graph structures. Existing methods address this challenge by employing a GNN-nested transformer, as seen in Graphformers~\cite{yang2021graphformers} to capture both textual attributes and their correlations between nodes, or by adopting structure-aware objectives to fine-tune LMs~\cite{chien2021node}. While the former approach is effective, it may encounter efficiency issues due to the combination of GNNs and Transformers. Conversely, the latter approach is computationally efficient, but it predominantly focuses on learning from individual TAGs, leaving joint learning from diverse TAGs relatively under-explored. 

To bridge the gap, we pursue the second direction by training a unified language model $f$ using structure-aware learning signals from multiple TAGs $\{G_i\}_{i=1}^m$. Specifically, let $v$ represent an arbitrary node across $m$ TAGs, and $\mathcal{T}_v$ denote its corresponding text attribute. The collaborative pre-training objective for node $v$ is defined as: 
\begin{equation}
\begin{split}
\mathcal{L}_v = -\sum_{u\in \mathcal{S}_v} \log \frac{\exp \left( {\text{sim}(f(\mathcal{T}_v), f(\mathcal{T}_u)})/{\tau} \right)}{\sum\limits_{k \in \mathcal{B}} \exp \left( {\text{sim}(f(\mathcal{T}_v), f(\mathcal{T}_k)})/{\tau} \right)},
\end{split}
\label{conloss}
\end{equation}
where $\mathcal{S}_v$ denotes the set of structurally similar nodes to $v$, $\mathcal{B}$ is the sampled batch set in mini-batch training with $v\in \mathcal{B}$, and $\tau$ signifies the temperature parameter. $\text{sim}(\cdot,\cdot)$ is a similarity function such as the inner product, and $f$ is a PLM encoder, enabling the encoding of text attributes into embedding vectors. By optimizing Eq.~\eqref{conloss}, the language encoder $f$ is trained to generate similar representations for node $v$ and nodes in $\mathcal{S}_v$, while simultaneously pushing away the representations of $v$ and other nodes in the mini-batch set $\mathcal{B}$. Notably, as nodes in $\mathcal{B}$ are randomly sampled across $m$ TAGs, Eq.~\eqref{conloss} provides a simple yet effective way to learn from nodes in various scenarios due to its instance-wise discriminator nature.

While conceptually simple and feasible, learning through Eq.~\eqref{conloss} faces two major challenges in practice. \textbf{C1: The structurally similar node set $\mathcal{S}_v$ of $v$ is not well-defined.} Given the heterogeneity of various TAGs and node-level statistics, random sampling based on neighbors may be suboptimal for capturing the diversity between nodes across different TAGs. \textbf{C2: It presents a trade-off between model performance and training efficiency.} The computational costs of Eq.~\eqref{conloss} are determined by the number of positive samples ($|\mathcal{S}_v|$) and batch size ($|\mathcal{B}|$), as it requires the language model to encode $(|\mathcal{S}_v|+1)|\mathcal{B}|$ text sequences per iteration. While reducing the number of positive samples can accelerate training, it may degrade performance since sufficient structural information is critical for graph contrastive learning~\cite{you2020graph,zhu2021graph,zhang2024graph}. In Section~\ref{sec-samples} and Section~\ref{sec-memory-efficiency}, we introduce two strategies to address these challenges, respectively.

\subsection{Adaptive and Learnable Positive Scheme}
\label{sec-samples}
To extract the set of similar structural nodes $\mathcal{S}_v$ of $v$ in Eq.~\eqref{conloss} (\textbf{C1}), the conventional protocol aims to randomly sample some nodes from the neighbors of $v$', that is, nodes directly connected in the original TAG. However, this seemingly intuitive approach is suboptimal within our learning scenarios, given that neighborhood distributions frequently exhibit significant variability both within and across graphs. To effectively consolidate critical structure information across diverse TAGs, we posit that an advanced sampling strategy should account for the following essential factors. 
\begin{itemize}[leftmargin=*]
\item \textbf{Local Structure.} Leveraging the local neighborhood structure, i.e., directly connected nodes, constitutes a fundamental design principle of GNNs~\cite{kipf2016semi}. To achieve a unified alignment of text attributes and graph structures across various TAGs, the local neighbors of nodes are indispensable. 
\item \textbf{High-Order Structure.} Beyond the local structure, high-order neighbors are crucial for the efficacy of graph machine learning~\cite{liu2020towards,li2021training}, particularly for long-tail nodes~\cite{liu2021tail}.
\item \textbf{Graph Statistics.} Unlike the standard learning paradigm, learning from multiple TAGs necessitates consideration of the unique characteristics inherent to each graph. For example, node degree distributions vary across TAGs, leading to diverse interpretations of the hub nodes. Additionally, node status are graph-specific and not directly comparable between TAGs.  
\end{itemize}

\subsubsection{Adaptive Positive Sampling} Motivated by these observations, we propose an innovative positive sample selection scheme that adaptively sample structurally similar neighbors by considering nodes' local and high-order neighbors, as well as their unique statuses within each graph.  The selection is weighted on the basis of the unique statistical characteristics of each graph. 
Given graph ${G}_i$ and a central node $v$, positive samples set $\mathcal{S}_v$ of $v$ is denoted as: $\mathcal{S}_v = \text{AdaPS}(v, {G}_i, t)$. We define function $\text{AdaPS()}$ as follows: 
\begin{equation}
\small
\text{AdaPS}(v, {G}_i, t) = 
\begin{cases} 
N_1(v) + N_h(v) & \text{if } \deg(v), \deg(\overline{{G}_i}) < t \\
N_1(v) & \text{otherwise}
\end{cases} .
\end{equation}

Here, $t$ is the number of positive samples for each central node. \(\deg(v)\) represents the degree of node $v$ and $\deg(\overline{{G}_i)}$ is the average degree of $G_i$. $N_1(v)$ and $N_h(v)$ denote the first-hop and high-order neighbor set of $v$, respectively. If number of nodes in $N_1(v)$ exceeds $t$, select the top $t$ nodes with highest Personalized Pagerank Score~\cite{page1999pagerank}.  For each central node, we select positive samples adaptively by considering both individual node statistics and overall graph metrics. This approach ensures that nodes with local or high-order structure information are chosen as candidates properly for each central node with minimal noise.

\textit{Remark:} Our design strategically incorporates both local and higher-order structural information through the candidate selection function \(C\), enhancing the depth and accuracy of structural insights in TAGs. By further integrating the Personalized PageRank scores in our adaptive sampling function \(AdaPS\), the model prioritizes key nodes based on their centrality, adapting effectively to each graph's unique statistical characteristics. This comprehensive approach ensures robust performance and superior adaptability across various graph topologies.

\subsubsection{Contrasting with Informative Positive Samples}
\label{sec-aggregation}
Given the sampled positive pool $S_v$ of node $v$, one can use Eq.~\eqref{conloss} to train the unified PLM across multiple TAGs. However, this approach maybe suboptimal since it treats all positive samples equally. As verified in previous studies~\cite{you2021graph,yin2022autogcl}, the generation of positive samples are rather important for contrastive learning on graphs. While we do not generate positive samples using explicit graph augmentation strategies, as done in previous efforts~\cite{you2020graph,zhang2024graph}, the adaptive positive generation procedure itself could be regarded as an intuitive implementation of data augmentation, identifying informative positive samples. Thus, the quality of positive samples in $S_v$ varies from node to node. In our learning scenario, this may be exacerbated by the inherent heterogeneity of graph types due to the diversity of domains. Therefore, a tailored contrastive objective is needed to boost the performance of our learning from multiple TAGs problem. 

To achieve this, we devise a learnable positive generation module, which adaptively selects the most informative positive samples from the pool for loss calculation, by leveraging the personalized node status and domain knowledge of anchor node and corresponding positive candidates. Specifically, let $\mathcal{T}^D_v$ denote the domain-specific context description of node $v$ including its status and domain knowledge (please refer to Appendix More Experiment Details Section for details), we can rewrite Eq.~\eqref{conloss} as follows: 

\begin{equation}
\begin{split}
\mathcal{L}_v = -\log \frac{\exp \left( \text{sim}(f(\mathcal{T}_v), \text{Pos}(\mathcal{T}_v, S_v, \mathcal{T}^D_v))) / \tau \right)}{\sum\limits_{k \in \mathcal{B}} \exp \left( \text{sim}(f(\mathcal{T}_v), f(\mathcal{T}_k)) / \tau \right)},
\end{split}
\label{conloss-new}
\end{equation}
where, $\text{Pos}(\cdot,\cdot,\cdot)$ is a positive generator, which generates an informative positive sample for node $v$ by using the textual attributes of positive candidates in $S_v$ and its domain knowledge $\mathcal{T}^D_v$. In particular, $\text{Pos}()$ adaptively selects the most informative positive candidates to constitute the final positive sample as follows.
\begin{equation}
\text{Pos}(\mathcal{T}_v, S_v, \mathcal{T}^D_v) = \sum_{u \in S_v} \frac{\exp (\mathbf{E}_{pos}^v(u))}{\sum_{u^\prime} \exp (\mathbf{E}_{pos}^v(u^\prime))}\cdot f(\mathcal{T}_u).
\label{equ-learnable}
\end{equation}
$\mathbf{E}_{pos}^v\in\mathbb{R}^{|S_v|}$ is the learnable selection table associated with node $v$, where $\mathbf{E}_{pos}^v(u))$ represents the importance of candidate $u$ to $v$. Specifically, $\mathbf{E}_{pos}^v(u))$ is initialized as: $\mathbf{E}_{pos}^v(u))=\text{sim}(f_0(\mathcal{T}_v + \mathcal{T}^D_v), \mathcal{T}_u)$, and $f_o$ stands for the original PLM chosen. By doing this, personalized node characteristics and graph scenarios are incorporated into our learning process. 



In the optimization process, we add a regularization term on $\mathbf{E}_{pos}^v$ to ensure that the learned importance distribution aligns in general well with the corresponding graph topological structure, expressed as:

\begin{equation}
\mathcal{L}_{KL}^v = \text{KL}(\text{softmax}(\mathbf{E}_{pos}^v), \text{softmax}(\text{Page}^v)).
\end{equation}
Here, $\text{Page}^v\in\mathbb{R}^{|S_v|}$ represents the importance vector of node $v$ w.r.t. different nodes in $S_v$, where the importance value is estimated by the Personalized PageRank algorithm~\cite{page1999pagerank}. $\text{KL}$ stands for the Kullback–Leibler divergence between two distributions. 

\textit{Remark:} Our learnable positive generation scheme provides an adaptive way for domain-specific contrastive learning objective. It aligns well with our learning over multiple TAGs scenarios since it can jointly take the graph statistics and node characteristics for positive generation, which excels in identifying the most informative positive sample for node contrasting, as verified in Figure~\ref{fig-learnable-distribution}. 

\subsection{The Lazy Contrastive Module}
\label{sec-memory-efficiency}
Another challenge in fine-tuning LMs from multiple TAGs using Eq.~\eqref{conloss} is the efficiency problem (\textbf{C2}). Given the constraints of GPU memory, there is a trade-off between the training batch size and the maximum number of positive samples considered. Increasing the batch size can accelerate the training speed by reducing the number of iterations per epoch, which is important given the large scale training nodes in our learning scenarios, yet at the expense of reducing the number of positive samples per node, and vice versa. However, as verified in previous studies~\cite{chien2021node,yi2024gaugllm}, preserving a sufficient number of structurally similar nodes is crucial to the success of graph contrastive learning.    

To address the dilemma, inspired by the momentum contrastive~\cite{he2020momentum}, we introduce a lazy contrastive module by treating positive sample encoding as dictionary look-up operation. Specifically, we establish a dynamic dictionary across various TAGs, which preserves and updates the representations of positive samples on-the-fly using nodes in the batch size, thereby avoiding the need to encode the text attributes of positive samples using LMs during the training. Formally, we rewrite the learnable positive generation process in Eq.~\eqref{equ-learnable} to an efficient version, expressed as:
\begin{equation}
\begin{split}
\text{Pos}(\mathcal{T}_v, S_v, \mathcal{T}^D_v) &= \sum_{u \in S_v} \frac{\exp (\mathbf{E}_{pos}^v(u))}{\sum_{u^\prime} \exp (\mathbf{E}_{pos}^v(u^\prime))}\cdot \mathbf{y}_u, \\
& s.t. \ \ \ \ \mathbf{y}_u = \text{LookUp}(\mathbf{E}, \text{Idx}(u)).
\end{split}
\label{efficient-loss}
\end{equation}

Here, $\text{LookUp}(\cdot,\cdot)$ denotes a simple embedding look-up operation based on the embedding table $\mathbf{E}^{n\times d}$ and the index of node $u$, where $n=\sum_i^m n_i$, $\text{Inx}(u)$ depends on both graph index and node index within the graph, and $d$ represents the hidden dimension. 
The overall loss for node $v$ is computed as \( \mathcal{L}_v=\mathcal{L}_{\text{vc}} + \alpha \mathcal{L}_{KL}^v\), where \( \alpha \) is a hyperparameter that adjusts the impact of the KL-divergence loss on the total loss.

Compared to Eq.~\eqref{equ-learnable}, learning through Eq.~\eqref{efficient-loss} is efficient, as the representations of positive samples $\mathbf{y}_*$ are obtained via embedding retrieval without the need for explicit text encoding. It is also worth noting that Eq.~\eqref{efficient-loss} is memory-efficient since $\mathbf{y}_*$ is gradient-free, reducing the abundance of intermediate tensors for gradient calculation. Consequently, a larger batch size can be used to further enhance the training speed. We empirically demonstrate the efficacy of these designs in Table~\ref{tab:time-cost}. 
Next, we will show how to effectively implement the lookup operation.

\noindent\textbf{Dictionary Update and Retrieval.} 
Given $m$ TAGs $\{\mathcal{G}_i\}_{i=1}^m$, we construct a dynamic embedding table $\mathbf{E}$ to store embedding of all nodes' text attributes in $m$ TAGs. Each node is uniquely identified in $\mathbf{E}$ by combining its own index within the graph and corresponding graph index. For example, let $v_{i,j}$ represent the $j$-th node in graph $G_i$, then its mapped node index in $\mathbf{E}$ is denoted as $\text{Idx}(v_{i,j})=j + \sum_{k=1}^{i-1} n_k$. Given $v_{i,j}$ and the intermediate LM encoder $f$, we can update $\mathbf{E}$ on-the-fly as follows. 
\begin{equation}
\begin{split}
\mathbf{E}(\text{Idx}(v_{i,j})) &= f(\mathcal{T}_{v_{i,j}}), \quad v_{(i,j)} \in \mathcal{B}.
\end{split}
\end{equation}
Here, only nodes in $\mathcal{B}$ are used to update the embedding table $\mathbf{E}$, which gives rise to the name "lazy", since it utilizes the encoded representations in previous iteration for dictionary updating. In parallel, given the index of positive sample $u_{i,j}$ of node $v_{i,j}$, we extract its hidden representation via simple indexing, i.e., $\mathbf{y}_{u_{i,j}}\mathbf{E}(Idx(u))$, which is LM encoding-free and can accelerate the training speed.

\textit{Remark:} In contrast to MoCo~\cite{he2020momentum}, we do not employ an additional momentum LM encoder to update the embedding table $\mathbf{E}$ over time. Instead, we directly utilize the encoded central nodes from previous training steps as the latest representations for updating $\mathbf{E}$. This design not only enhances our training speed, as demonstrated in Table~\ref{tab:time-cost}, but also results in notable performance improvements, as empirically verified in Table~\ref{table-nc-main}, as it encourages the LM encoder to learn from previous experiences.
\begin{table}[ht!]
 \centering
 \small 
 \begin{tabular}{lcccc}
 \toprule
 & GIANT & PATTON & UniGLM w/o E & UniGLM \\
 \midrule
 Training time & 14.24h & 5.4d & 63h & 13.5h \\
 Inference time & 25min & 5h15min & 20min & 20min \\
 \bottomrule
 \end{tabular}
 \caption{Training and inference efficiency comparison. For GIANT and PATTON, training time is the sum of time training individual on seven datasets. For UniGLM, it is the time used for training one model with all datasets. For inference time, the time it needs to encode ogbn-arxiv is reported.}
 \label{tab:time-cost}
\end{table}

\begin{table*}[h!]
\centering
\resizebox{\textwidth}{!}{
\begin{tabular}{c|c|c|c|c|c} 
\toprule
Dataset & Emb Types & MLP & GCN & SAGE & RevGAT \\ 

\midrule
\multirow{6}{*}{Computers} & SE & 57.34±0.47 (+33.24\%)& 70.22±0.40 (+17.03\%)& 71.02±0.27 (+16.73\%)& 70.54±0.23(+17.28\%)\\
 & BERT & 54.04±0.20 (+41.38\%)& 66.88±0.24 (+22.88\%)& 67.25±0.12 (+23.27\%)& 65.69±0.05(+25.94\%)\\
 & GIANT & \underline{73.05±0.31 (+4.59\%)}& \underline{80.09±0.16 (+2.61\%)}& \underline{81.03±0.12 (+2.31\%)}& \underline{80.99±0.18(+2.15\%)}\\
 & PATTON & 71.60±0.30 (+6.70\%)& 78.64±0.14 (+4.50\%)& 79.98±0.15 (+3.65\%)& 78.97±0.23(+ 4.76\%)\\
 & MixGIA & 68.83±0.21 (+11.00\%)& 77.17±0.11 (+6.49\%)& 76.96±0.26 (+7.72\%)& 77.22±0.42(+7.14\%)\\
 & UniGLM & \textbf{76.40±0.11} & \textbf{82.18±0.08} & \textbf{82.90±0.11} & \textbf{82.73 ± 0.19}\\

\midrule
\multirow{6}{*}{Ogbn-Arxiv} & SE & 64.30±0.09 (+15.61\%) & 71.74±0.29 (+2.82\%) & 71.49±0.27 (+5.50\%) & 74.02±0.18(+1.76\%)\\
 & BERT & 66.29±0.20 (+12.14\%) & 72.63±0.31 (+1.56\%) & 73.33±0.33 (+2.85\%) & 72.88±0.39(+3.35\%)\\
 & GIANT & 73.08±0.06 (+1.72\%) & 73.29±0.10 (+0.64\%) & 74.59±0.28 (+1.11\%) & \textbf{75.96±0.09(--0.84\%) }\\
 & PATTON & \underline{73.47±0.11 (+1.18\%)}& \textbf{73.59±0.20 (+0.23\%)}& \underline{75.00±0.16 (+0.56\%)}& 74.08±0.12(+1.67\%)\\
 & MixGIA & 69.30±0.18 (+7.27\%) & 72.94±0.19 (+1.12\%) & 73.64±0.20 (+2.42\%) & 72.39±0.51(+4.05\%)\\
 & UniGLM & \textbf{74.34±0.09} & \textbf{73.76±0.12} & \textbf{75.42±0.21} & \underline{75.32±0.22}\\
 \midrule
\multirow{6}{*}{Ogbn-Products} & SE & 53.85±0.17 (+43.83\%) & 70.52±0.51 (+10.55\%) & 69.13±0.26 (+13.57\%) & 69.64±0.17(+13.64\%)\\
 & BERT & 67.58±0.28 (+14.60\%) & 74.77±0.87 (+4.27\%) & 74.09±0.27 (+5.97\%) & 74.53±0.26(+6.19\%)\\
 & GIANT & 72.46±0.33 (+6.89\%) & 69.77±0.42 (+11.74\%) & 68.69±1.19 (+14.30\%) & 71.89±0.30(+10.08\%)\\
 & PATTON & \underline{76.42±0.23 (+1.35\%)}& \underline{77.22±0.34 (+0.96\%)}& \underline{77.81±0.58 (+0.90\%)}& \underline{78.48±0.10(+0.84\%)}\\
 & MixGIA & 70.44±1.54 (+9.95\%) & 75.73±0.57 (+2.94\%) & 75.90±0.84 (+3.44\%) & 74.88±1.19(+5.69\%)\\
 & UniGLM & \textbf{77.45±0.10} & \textbf{77.96±0.22} & \textbf{78.51±0.33} & \textbf{79.14±0.22} \\
\midrule
\multirow{6}{*}{Overall} & SE & 60.98±0.24 (+24.22\%) & 69.66±0.42 (+10.44\%) & 69.98±0.37 (+10.90\%) & 70.13±0.35(+10.77\%)\\
 & BERT & 63.66±0.31 (+18.99\%) & 70.27±0.45 (+9.48\%) & 70.69±0.40 (+9.79\%) & 70.22±0.37(+10.62\%)\\
 & GIANT & 73.62±0.28 (+2.89\%) & 74.56±0.33 (+3.18\%) & 75.61±0.35 (+2.65\%) & 76.02±0.34(+2.18\%)\\
 & PATTON & \underline{74.12±0.26 (+2.20\%)}& \underline{75.75±0.31 (+1.56\%)}& \underline{76.80±0.30 (+1.05\%)}& \underline{76.56±0.29}(+1.46\%)\\
 & MixGIA & 70.80±0.30 (+6.99\%) & 74.60±0.34 (+3.12\%) & 74.92±0.32 (+3.59\%) & 74.42±0.33(+4.38\%)\\
 & UniGLM & \textbf{75.75±0.25} & \textbf{76.93±0.28} & \textbf{77.61±0.27} & \textbf{77.68±0.26}\\
\midrule
\bottomrule
\end{tabular}
}
\caption{Semi-supervised accuracy results on MLP and state-of-art GNNs with various embeddings for History, Computers, Fitness and Ogbn-Arxiv datasets. Top 2 embedding types are bold or underlined. If gap is less than 0.5\%, we consider it as both Top 1. More results can be seen in Appendix.}
\label{table-nc-main}
\end{table*}

\section{Experiments}
\label{experiments}
We aim to answer the following research questions. \textbf{RQ1}: How does UniGLM perform against leading graph embedding models in terms of node classification and link prediction tasks?
\textbf{RQ2}: How well does UniGLM transfer in cross-domain and in-domain scenarios? 
\textbf{RQ3}: How does each component of UniGLM, i.e., sampling strategy, learnable generation and efficient embedding table, contribute to the performance? 
\textbf{RQ4}: What is the impact of different pre-trained language model backbones on UniGLM?

\subsection{Experiments Setup}
We train UniGLM on seven TAG datasets (details in Appendix) from two major domains: academic networks and Amazon products. UniGLM is compared with multiple embedding models for node classification and link prediction. Node classification is conducted with three different types of downstream backbones: MLP, GNNs (GCN~\cite{kipf2016semi}, GraphSAGE~\cite{hamilton2017inductive}, and RevGAT~\cite{li2021training}), and self-supervised graph methods (GraphCL~\cite{you2020graph} and S2GAE~\cite{tan2023s2gae}), while link prediction uses MLP. Our experiments focus on semi-supervised and transfer learning settings. All baselines are previously discussed in Section \ref{related-work}. Additionally, MixGIANT is implemented by us as a variant of GIANT~\cite{chien2021node} that can be trained on multiple datasets. More details can be found in the Appendix.

\subsection{Performance Comparison}
\label{semisupervised-experiments}
\textbf{Node Classification.} To address \textbf{RQ1}, we perform extensive experiments on seven benchmark TAG datasets for node classification. Specifically, we first pre-train UniGLM on these datasets and then use the resultant model to generate node embeddings, which we then use as input features for downstream node classification tasks. The results are shown in Table~\ref{table-nc-main} and Table~\ref{table-nc-gssl}. Comprehensive results are in Appendix. We made the following observations: 

\textbf{\ding{172} UniGLM significantly outperforms existing graph embedding models in node classification task.} Table~\ref{table-nc-main} demonstrates that UniGLM consistently achieves superior performance across most datasets and backbone models, ranking first in 27 out of 28 cases. Although GIANT and PATTON improve over the SE method by leveraging language models and graph structures, UniGLM excels them in most cases. UniGLM's improvement over GIANT and PATTON indicates that leveraging structural information between graphs is also crucial, further emphasizing the importance of capturing and utilizing these relationships for enhanced performance. \textbf{\ding{173} UniGLM can enhance the performance across different backbones, demonstrating its wide applications.} By integrating both structural information within and between graphs, UniGLM can provide a stronger embedding for downstream tasks on a wide range of models.

\begin{table}[h!]
\centering
\begin{tabular}{c|c|c|c|c} 
\toprule
Dataset                  & Emb Types & Photo      & History    & Products  \\ 
\midrule
\multirow{6}{*}{GraphCL} & SE        & 63.34±0.27 & 76.14±0.11 & 64.12±0.05     \\ 
\cmidrule{2-5}
                         & BERT      & 55.18±0.89 & 79.55±0.26 & 67.52±0.31     \\ 
\cmidrule{2-5}
                         & GIA       & 58.39±0.32 & 68.24±0.09 & 73.21±0.15     \\ 
\cmidrule{2-5}
                         & PATTON    & 75.98±0.14 & \textbf{83.22±0.07} & 75.91±0.05     \\ 
\cmidrule{2-5}   
                         & UniGLM    & \textbf{78.55±0.10} & \textbf{83.22±0.06} & \textbf{77.08±0.06}     \\
\bottomrule
\end{tabular}
\caption{Semi-supervised accuracy results on classic graph contrastive learning method GraphCL with various embeddings for Photo, History, and Ogbn-Products datasets.}
\label{table-nc-gssl}
\end{table}

\textbf{Link Prediction.} To address \textbf{RQ1} for the link prediction task, we train UniGLM on seven datasets with edges in test set masked, using UniGLM as the encoder for the textual attributes of the node, and employ an MLP for downstream evaluation. Detailed split can be found in appendix. The results are illustrated in Figure~\ref{fig-lp-auc} and more resuls can be found in Appendix. We observe that \textbf{\ding{174}UniGLM exhibits a strong ability in link prediction, surpassing other baselines in most cases.} This means that the contrastive learning approach effectively captures and transfers relational patterns between nodes, thereby providing a robust model for various graphs.

\begin{figure}[h!]
    \centering
    \includegraphics[width=1.0\linewidth]{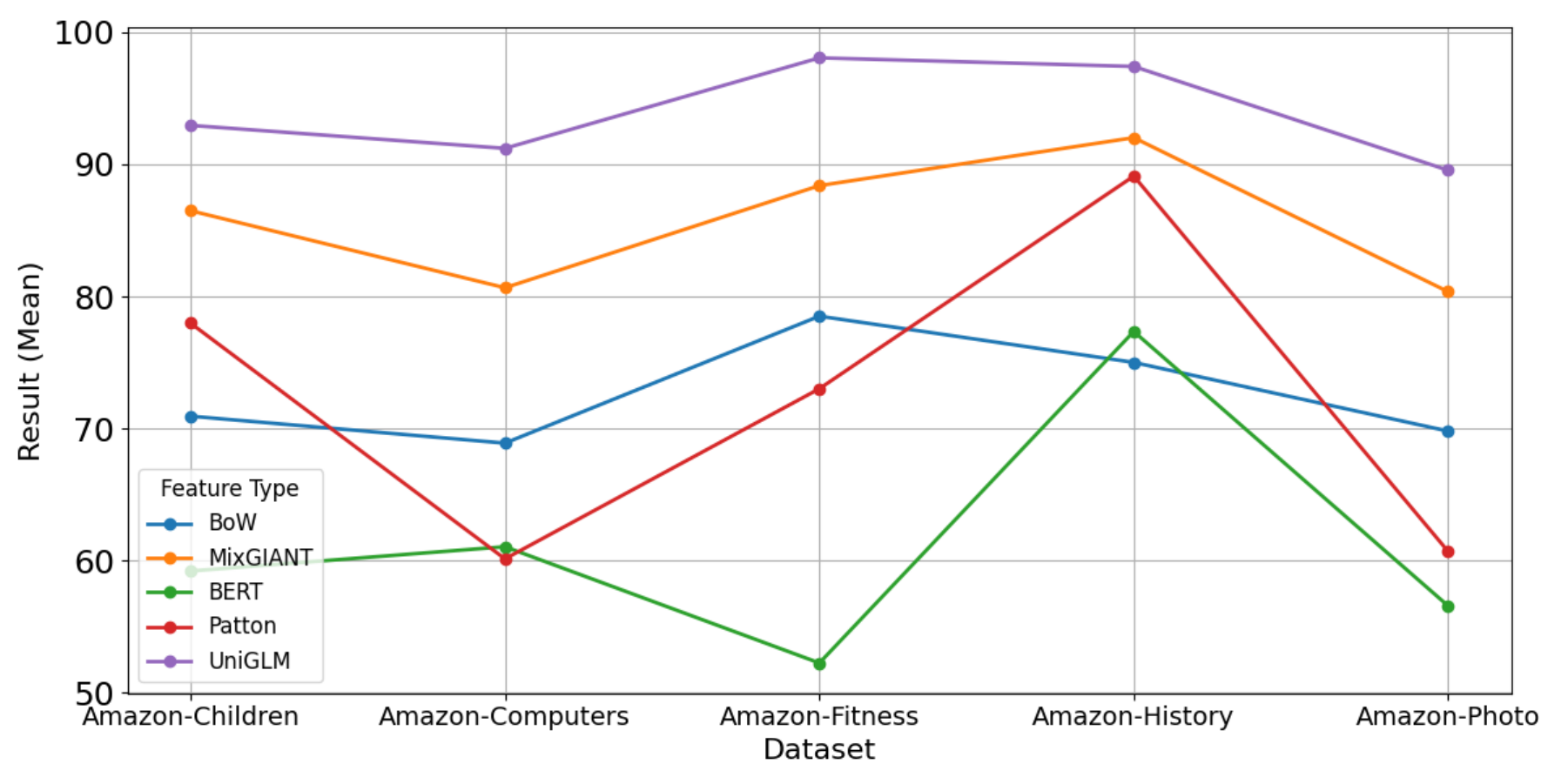}
    \caption{Link prediction results in AUC metric.}
    \label{fig-lp-auc}
\end{figure}

\subsection{Transfer Ability}
To answer \textbf{RQ2}, we conduct experiments from two perspectives: in-domain transfer and cross-domain transfer. 
\textbf{To assess in-domain transferability}, we use the \textbf{VideoGames} co-purchase network as a test case. Detailed settings can be found in the Appendix. Table \ref{tab-in-domain-transfer} suggest that \textbf{\ding{175} UniGLM demonstrate strong in-domain transfer ability in both tasks}. The results indicate that incorporating a diverse range of datasets can further enhance the model's generalization capability.

\begin{table}[h!]
\centering
\small 
\setlength{\tabcolsep}{3pt} 
\begin{tabular}{ccccccc}
\toprule
\multicolumn{7}{c}{\textbf{Link Prediction}} \\
\midrule
 &  & SE & BERT & MixGIANT & PATTON & UniGLM \\
\midrule
 & AP & 68.10±0.12 & 69.01±0.09 & 76.82±0.15 & 75.32±0.12 & \textbf{81.46±0.08} \\
 & AUC & 71.13±0.05 & 67.51±0.12 & 75.10±0.09 & 74.87±0.04 & \textbf{79.58±0.11} \\ 
 & Hits@100 & 0.72±0.04 & 2.01±0.02 & 4.04±0.09 & 4.12±0.12 & \textbf{9.03±0.08} \\
\midrule
\multicolumn{7}{c}{\textbf{Node Classification}} \\
\midrule
 &  & SE & BERT & MixGIANT & PATTON & UniGLM \\
\midrule
 & MLP & 40.55±0.68 & 41.88±0.45 & 45.59±0.36 & 47.39±0.64 & \textbf{48.70±0.67} \\
 & GCN & 49.28±0.72 & 49.09±0.54 & 49.76±0.37 & 50.16±0.51 & \textbf{51.91±0.63} \\
 & SAGE & 47.20±0.46 & 48.78±0.56 & 49.22±0.43 & 50.32±0.26 & \textbf{52.04±0.43} \\
\bottomrule
\end{tabular}
\caption{In-domain transfer ability comparison of UniGLM and baselines on unseen \textbf{VideoGames} dataset for link prediction and node classification tasks.}
\label{tab-in-domain-transfer}
\end{table}

\textbf{To assess the cross-domain transfer ability,} we use classic ogbn-arxiv as a test case. Detailed settings are reported in the Appendix and Table \ref{tab-cross-domain-transfer} lists partial results, from which we observe that \textbf{\ding{176}UniGLM presents great cross-domain transfer ability, surpassing all baselines in both semi-supervised and few-shot setting.} This findings further suggest that language model encoder can learn the structural pattern across a variety of datasets, achieving universal graph embedding.

\begin{table}[h!]
\centering
\small
\begin{tabular}{lcc|c}
\toprule
\multirow{2}{*}{Dataset} & \multirow{2}{*}{Method} & \multicolumn{2}{c}{Accuracy} \\
\cmidrule(r){3-4}
 & & Semi-Supervised & Few-Shot \\
\midrule
\multirow{6}{*}{Ogbn-Arxiv} 
& SE & 74.02±0.18 & 15.59±0.83 \\
& BERT & 66.29±0.20 & 14.72±1.13 \\
& MixGIANT & 72.44 ± 0.20 & 20.16±0.87 \\
& PATTON & 72.59 ± 0.12 & 21.08±1.19 \\
& UniGraph & 72.91±0.42 & 31.35±1.01 \\
& UniGLM & \textbf{74.60 ± 0.23} & \textbf{33.88±0.87} \\
\bottomrule
\end{tabular}
\caption{Cross-Domain evaluation of Ogbn-Arxiv under semi-supervised and few-Shot settings.}
\label{tab-cross-domain-transfer}
\end{table}

\subsection{Ablation Study}
To answer \textbf{RQ3}, we evaluate the performance of the UniGLM after removing or replacing some key components.

\begin{figure}[h!]
    \centering
    \includegraphics[width=0.9\linewidth]{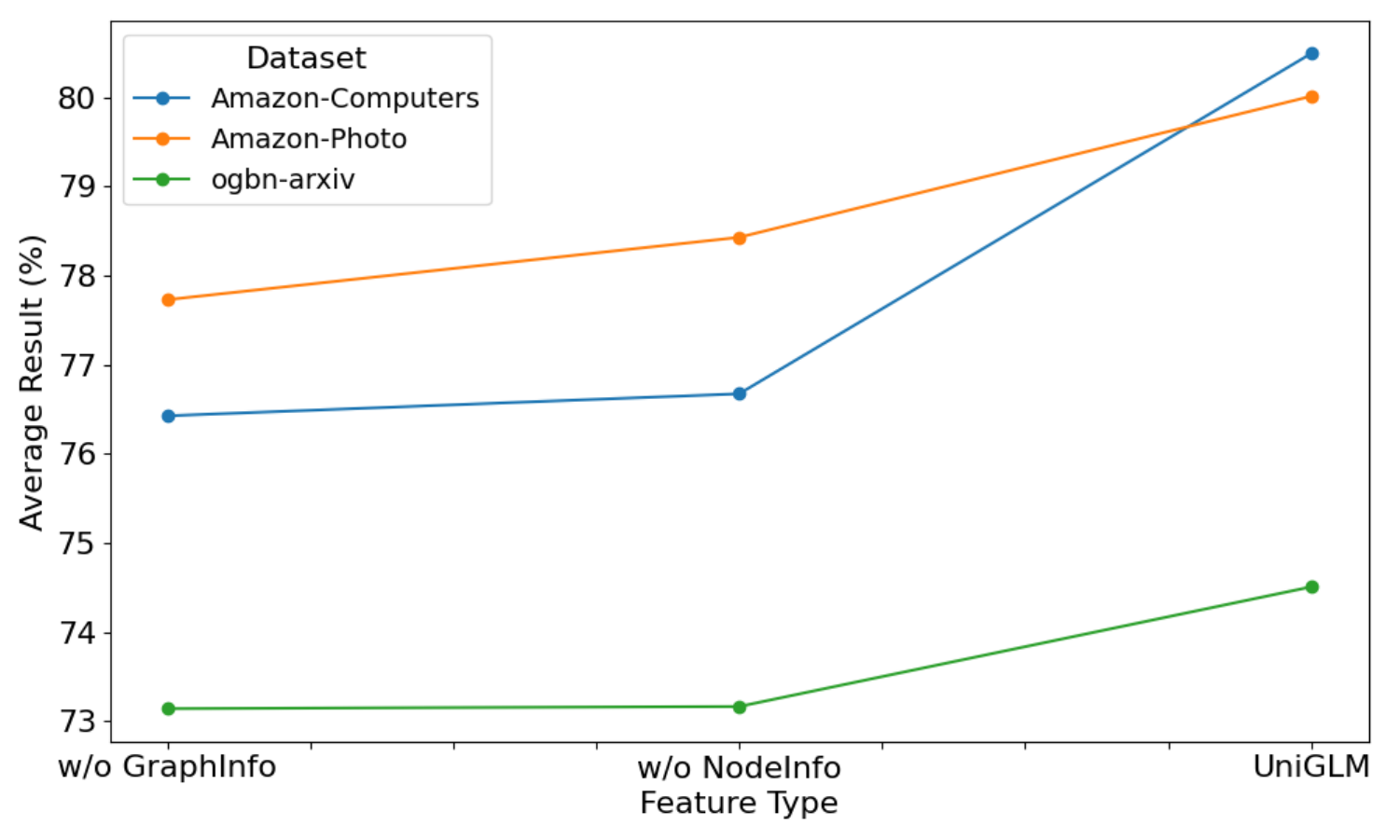}
    \caption{Ablation Study: the impact of different sample strategies. Results are the average of MLP, GCN and SAGE backbones.}
    \label{fig-ablation-sample}
\end{figure}

\textbf{The impact of sampling strategy.} We designed two variants: one without graph information named ``GraphInfo'', where the average degree of the graph is not considered during sampling, and the other is without node information named ``NodeInfo'', where the degree of individual nodes is not considered. From Figure~\ref{fig-ablation-sample}, we observe that \textbf{\ding{177} Both node degree and graph density information can be used to boost the performance of UniGLM.} The possible reason is that domain knowledge is useful for our model to align structure information from various domains in a personalized fashion. 

\begin{figure}
    \centering
    \includegraphics[width=0.9\linewidth]{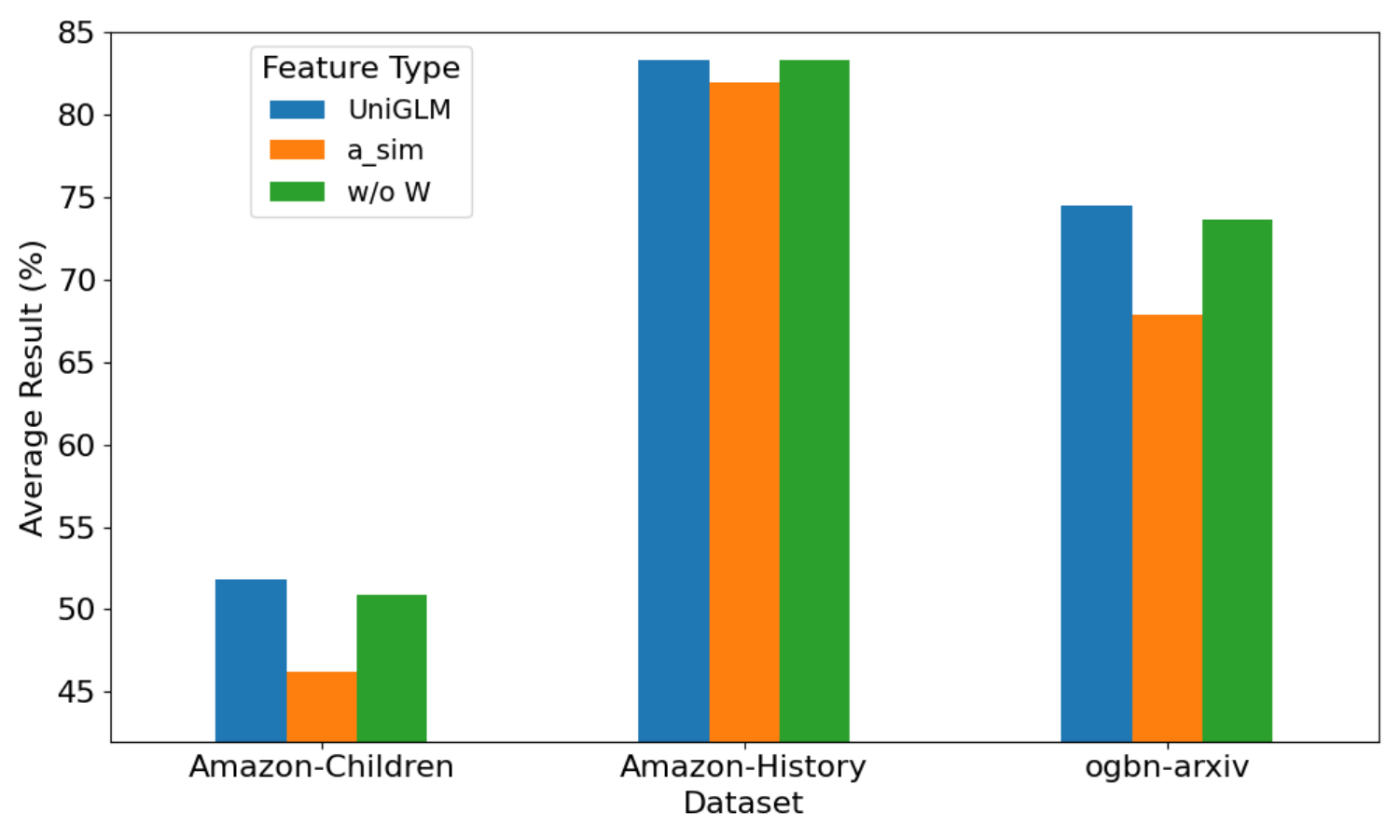}
    \caption{The impact of learnable positive generation scheme on UniGLM. The results are averaged values across three backbones: MLP, GCN, and SAGE. 
    }
    \label{fig-ablation-aggregation}
\end{figure}

\begin{figure}
    \centering
    \includegraphics[width=1.0\linewidth]{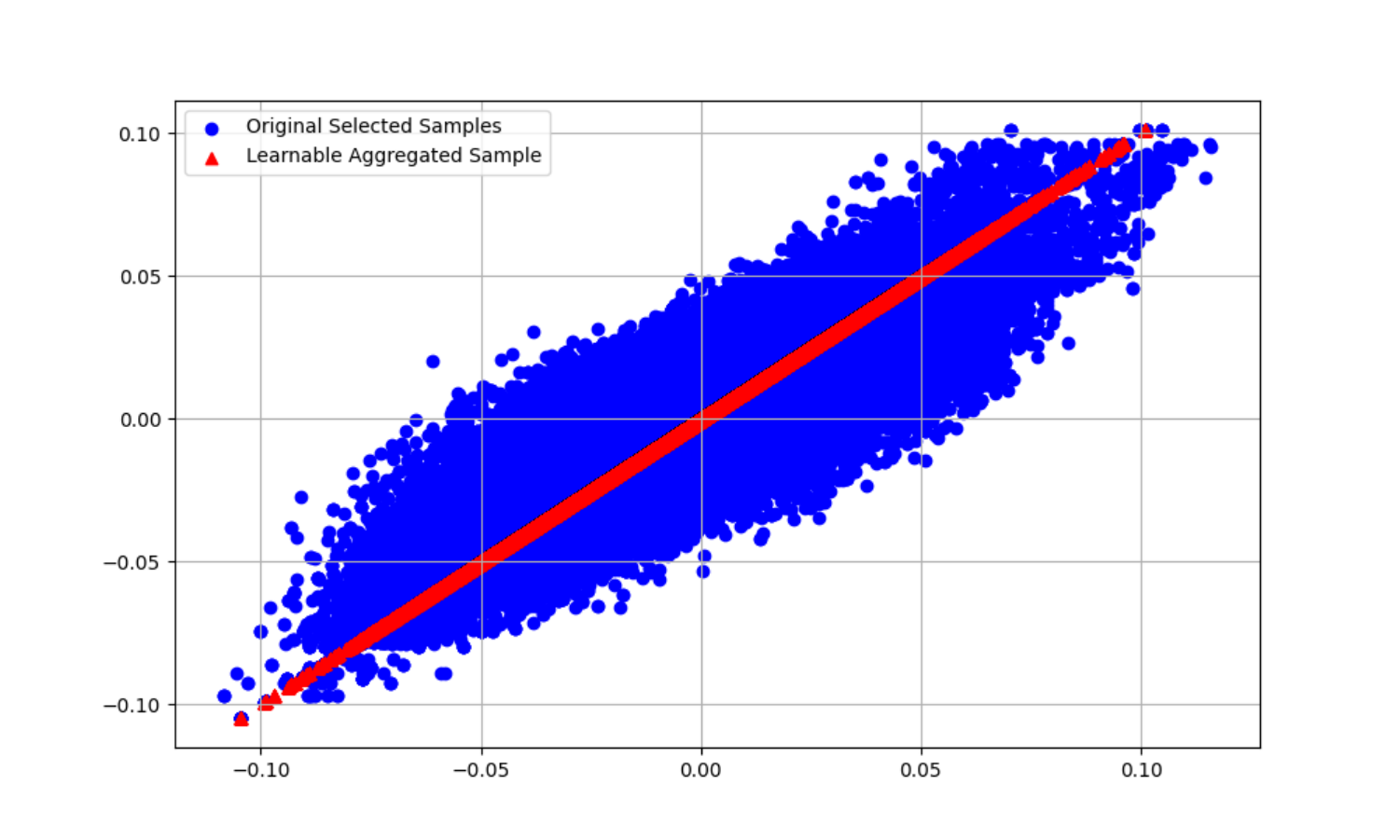}
    \caption{Visualization of the generated positive sample and the corresponding positive candidates on History dataset.}
    \label{fig-learnable-distribution}
\end{figure}

\textbf{The imapct of learnable positive generation module.} We introduce two variants, ``w/o W", which fixes the weight instead of make it learnable. The second approach is ``a\_sim", which uses the weights to aggregate the similarities between the positive samples and the center node, rather than generating the embeddings of the positive samples themselves before computing similarity. The results are shown in Figure~\ref{fig-ablation-aggregation}, from which we observe that \textbf{\ding{178} Our learnable positive generation module performs consistently better than other variants.} We attribute this improvement to the selection mechanism inherent in the positive embedding generation process, which has the potential to constitute more informative positive samples. To further verify this, we visualize the positive candidates and the generated positive embedding in Figure~\ref{fig-learnable-distribution}. We observe that the generated positive samples often lie in the middle of corresponding positive candidates, leading to an informative positive sample for constrastive loss compared with original ones. 

\begin{figure}
    \centering
    \includegraphics[width=0.9\linewidth]{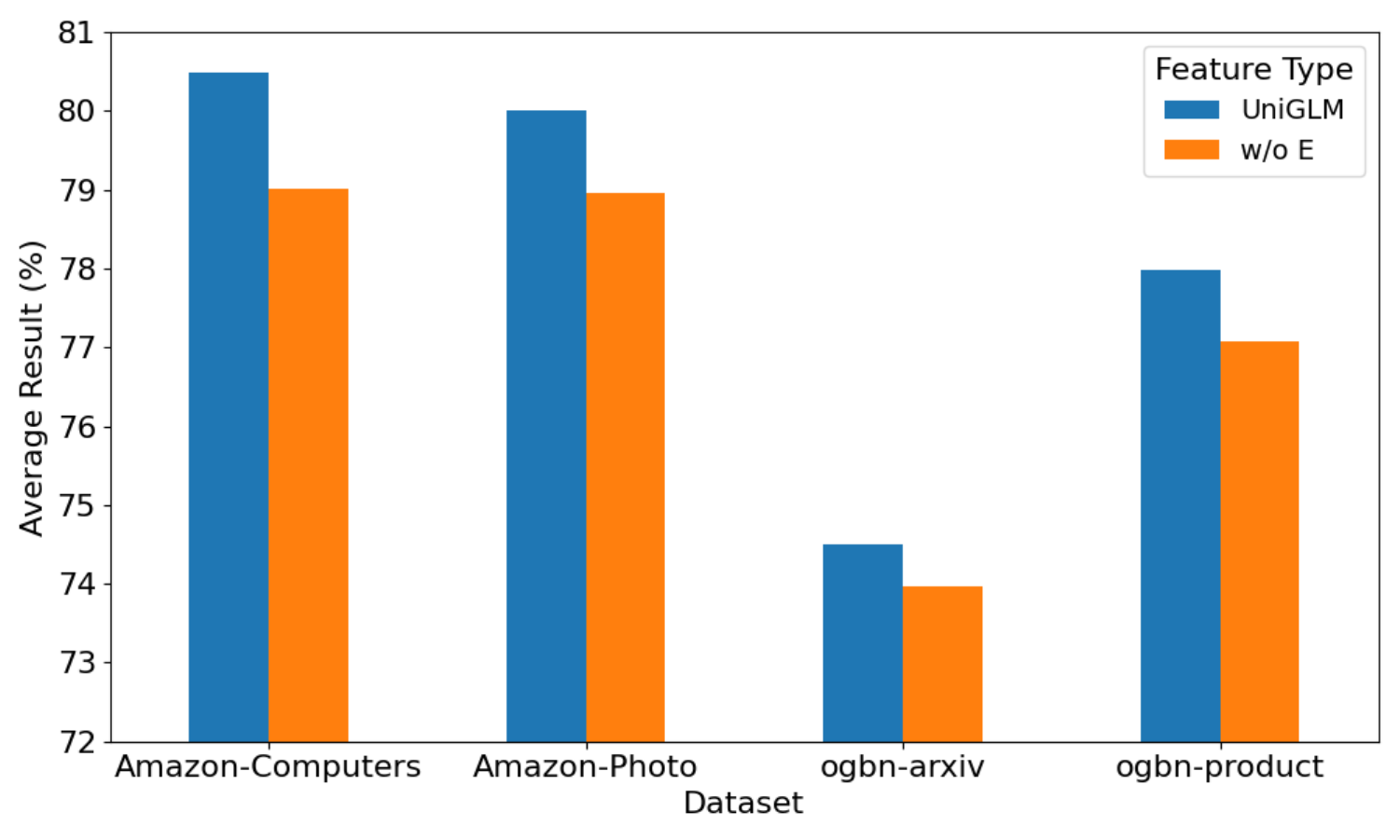}
    \caption{The impact of lazy contrastive module on UniGLM. Results are the average of MLP, GCN and SAGE backbones.}
    \label{fig-ablation-EmbeddingTable}
\end{figure}

\textbf{How effective is the proposed embedding table module?} We study the influence of the lazy contrastive module in Section 4.3. From the efficiency perspective, as shown in Table~\ref{tab:time-cost}, \textbf{\ding{179} Our lazy contrastive module can significantly accelerate the training speed}, since it avoids the calculations of positive samples on-the-fly. More interestingly, we empirically observe that this lazy updating does not impair the model performance as expected. In contrast, it can also boost our model performance as shown in Figure~\ref{fig-ablation-EmbeddingTable}. The feasible reason is that this lazy update rule allows our model to remember the most recent record during training, thus greatly reducing the risk of deviating from the previous step based solely on the current mini-batch.  

To answer \textbf{RQ4}, we test the performance of UniGLM under three different PLM backbones: MPNET\cite{song2020mpnet}, BERT and DeBERTa\cite{he2020deberta}. The results are summarized in Table \ref{fig-ablation-LMbackbone}, from which we observe that 
\textbf{\ding{180} The performance of UniGLM varies across various PLM backbones, yet it can achieve consistently better results using Sentence-BERT.}

\begin{table}[htbp]
\centering
\begin{tabular}{ccccc}
\toprule
\textbf{Model} & \textbf{ogbn-arxiv} & \textbf{Children} & \textbf{Computers} \\
\midrule
Sentence-mpnet & 74.34 ± 0.09 & 52.36 ± 0.26 & 76.40 ± 0.11 \\
BERT           & 73.51 ± 0.11 & 51.92 ± 0.28 & 75.43 ± 0.08 \\
DeBERTa        & 73.23 ± 0.23 & 51.98 ± 0.42 & 74.82 ± 0.16 \\
\bottomrule
\end{tabular}
\caption{The impact of PLM backbones on UniGLM.}
\label{fig-ablation-LMbackbone}
\end{table}

\section{Conclusion}
In this paper, we introduce UniGLM, a unified graph embedding framework aimed at fine-tuning pre-trained language model on a set of TAG datasets from various domains. UniGLM features two main innovations: (1) an adaptive and learnable positive scheme that selects positive samples using node's local, global, and domain knowledge, which is subsequently used for informative positive sample generation, and (2) a dynamic embedding table that efficiently encodes these samples on-the-fly to speed up training. We validate UniGLM across diverse TAGs from different domains, where it consistently surpasses existing methods in node classification, link prediction, demonstrating its superior transfer ability, effectiveness and efficiency.




\bibliographystyle{ACM-Reference-Format}
\bibliography{reference}










\end{document}